\documentclass{article}
\usepackage{spconf,amsmath,epsfig}

\usepackage{pstricks}
\usepackage{tikz}
\usetikzlibrary{calc}
\usepackage{amsmath}
\usepackage{lineno}
\usepackage{amssymb}
\usepackage{graphicx}
\usepackage{tumabbrev}
\usepackage{marginnote}
\usepackage{csquotes}
\usepackage[nohyperlinks]{acronym}
\usepackage[xindy,acronym,nohypertypes=acronym]{glossaries}
\defglsdisplayfirst[acronym]{\emph{#1#4}}

\newcommand{\WNetcGAN}{\textsc{WNet-cGAN}} 
 
\newcommand{\WNet}{\textsc{WNet}}

\usepackage[capitalize]{cleveref}

\usepackage{fancyhdr}
\usepackage{subcaption}
\usepackage{siunitx}
\usepackage[
  backend=bibtex,
  bibstyle=ieee,
  citestyle=numeric-comp,
  natbib=true,
  mincitenames=1,
  maxcitenames=1,
]{biblatex}
\bibliography{refs}

\DeclareSIUnit\px{px}

\graphicspath{{./figures/}}

\newacronym{gl:DSM}{DSM}{digital surface model}
\newacronym{gl:nDSM}{nDSM}{normalized digital surface model}
\newacronym{gl:DTM}{DTM}{digital terrain  model}
\newacronym{gl:LIDAR}{LIDAR}{light detection and ranging}
\newacronym{gl:GAN}{GAN}{generative adversarial network}
\newacronym{gl:cGAN}{cGAN}{conditional generative adversarial network}
\newacronym{gl:cLSGAN}{cLSGAN}{conditional least square generative adversarial network}
\newacronym{gl:LoD}{LoD}{level of details}
\newacronym{gl:CityGML}{CityGML}{city geography markup language}
\newacronym{gl:SGM}{SGM}{semi-global matching}
\newacronym{gl:SGD}{SGD}{stochastic gradient descent}
\newacronym{gl:OSM}{OSM}{open street map}
\newacronym{gl:RMSE}{RMSE}{root mean squared error}
\newacronym{gl:MAE}{MAE}{mean absolute error}
\newacronym{gl:NMAD}{NMAD}{normalized median absolute deviation}
\newacronym{gl:NCC}{NCC}{normalized correlation coefficient}
\newacronym{gl:FCN}{FCN}{fully convolutional Network}
\newacronym{gl:MBR}{MBR}{minimum boundary rectangles}
\newacronym{gl:RANSAC}{RANSAC}{random sample consensus}
\newacronym{gl:GSD}{GSD}{ground sampling distance}
\newacronym{gl:RJMCMC}{RJMCMC}{recursive jump monte carlo markov chain}
\newacronym{gl:SA}{SA}{simulated annealing}
\newacronym{gl:CNN}{CNN}{convolutional neural network} 
\newacronym{gl:CRF}{CRF}{conditional random field}
\newacronym{gl:InSAR}{InSAR}{interferometric synthetic aperture radar}
\newacronym{gl:SAR}{SAR}{synthetic aperture radar}
\newacronym{gl:GIS}{GIS}{geographic information systems}
\newacronym{gl:PAN}{PAN}{panchromatic}
\newacronym{gl:ReLU}{ReLU}{rectified linear units}
\title{DSM Building Shape Refinement from Combined Remote Sensing Images based on Wnet-cGANs }
\name{Ksenia Bittner\textsuperscript{1},Marco K\"orner\textsuperscript{2}, Peter Reinartz\textsuperscript{1}}
\address{\textsuperscript{1 }Remote Sensing Technology Institute, German Aerospace Center (DLR), Wessling, Germany - \\ 
	(ksenia.bittner, peter.reinartz)@dlr.de\\
	\textsuperscript{2 }Technical University of Munich, Munich, Germany - marco.koerner@tum.de\\}
\begin{document}
%\ninept
%
%{\let\newpage\relax\maketitle}
\maketitle
\begin{abstract}
%This article describes the workflow of a \glspl{gl:DSM} refinement algorithm using a hybrid \gls{gl:cGAN} where the generative part consists of two parallel networks merged at last stage forming a \WNet \space architecture.
We describe the workflow of a \glspl{gl:DSM} refinement algorithm using a hybrid \gls{gl:cGAN} where the generative part consists of two parallel networks merged at the last stage forming a \WNet \space architecture. 
The inputs to the so-called \WNetcGAN \space are stereo \glspl{gl:DSM} and \gls{gl:PAN} half-meter resolution satellite images.
%The fusion helps to propagate fine detailed information from a spectral image and complete the missing 3D knowledge from stereo \gls{gl:DSM} about building shapes. 
Fusing these helps to propagate fine detailed information from a spectral image and complete the missing 3D knowledge from a stereo \gls{gl:DSM} about building shapes. 
Besides, it refines the building outlines and edges making them more rectangular and sharp.
\end{abstract}
\glsresetall

\section{Introduction}
\label{sec:intro}

A \gls{gl:DSM} is an important and valuable data source for many remote sensing applications, like building detection and reconstruction, cartographic analysis, urban planning, environmental investigations and disaster assessment tasks.
The use of \gls{gl:DSM} for those remote sensing applications is motivated by the fact that it already provides geometric descriptions about the topographic surface.
With recent advances in sensor technologies, it became possible to generated \glspl{gl:DSM} with a \gls{gl:GSD} smaller than \SI{1}{\meter} not only from land surveying, aerial images, laser ranging data, or \gls{gl:InSAR}, but also using satellite stereo images.
The main advantages of satellite photogrammetric \glspl{gl:DSM} are the large land coverage and possibility to access remote areas. 
%Although, they provide good quality \glspl{gl:DSM}, they are not able to compete with large land coverage and low costs which the \glspl{gl:DSM} from satellite imagery reveal. 
However, \glspl{gl:DSM} generated with the image-based matching approaches miss objects like steep walls in urban areas or feature some unwanted outliers and noise due to temporal changes, matching errors or occlusions.
To overcome these problems, algorithms from computer vision have been analyzed and adapted to satellite imagery. 
For example, the filtering techniques such as geostatistical filter integrated with a hierarchical surface fitting technique, a threshold slope-based filter, or a Gaussian noise removal filter are the ones commonly used for \gls{gl:DSM} quality improvements.
Moreover, some methodologies propose to fuse \glspl{gl:DSM} obtained from different data sources to compensate the limitations and gaps which each of them has individually \cite{karkee2008improving}. 

With recent developments devoted to deep learning, it became possible to achieve top scores on many tasks including image processing. As a result, several works have already investigated their applicability for remote sensing applications, like landscape classification, building and road extraction, or traffic monitoring. 
Recently, a class of neural networks called \glspl{gl:GAN} was applied on three-dimensional remote sensing data and proved to be suitable.
Mainly, the generation of large-scale 3D surface models with refined building shape to the \gls{gl:LoD} 2 from stereo satellite \glspl{gl:DSM} was studied using \glspl{gl:cGAN}~\cite{bittner2018automatic,bittner2018dsm}. 
In this paper, we follow those ideas and propose a hybrid \gls{gl:cGAN} architecture which couples half-meter resolution satellite \gls{gl:PAN} images and \glspl{gl:DSM} to produce 3D surface models not only with refined 3D building shapes, but also with their completed structures, more accurate outlines, and sharper edges.  

%% Some notation
% \newcommand{}{\ensuremath{}\xspace}

\newcommand{\generator}{\ensuremath{G}\xspace}
\newcommand{\discriminator}{\ensuremath{D}\xspace}
\newcommand{\e}{\ensuremath{\mathrm{e}}}

\section{Methodology}
\label{sec:Methodology}

%The birth of \gls{gl:GAN}-based domain adaptation neural networks introduced by \citet{goodfellow2014generative} has made great achievements in generating realistic images.
The birth of \gls{gl:GAN}-based domain adaptation neural networks introduced by \citet{goodfellow2014generative} yielded great achievements in generating realistic images.
The idea behind the adversarial manner of learning is to train a pair of networks in a competing way: a \emph{generator} $\generator$ that tries to fool the discriminator to make the source domain look like the target domain as much as possible, and a \emph{discriminator} $\discriminator$ that tries to differentiate between the target domain and the transformed source domain. 
Taking the source distribution as input instead of a uniform distribution and using this external information to restrict both the generator in its output and the discriminator in its expected input leads to the conditional type of \glspl{gl:GAN}. 
The objective function for \glspl{gl:cGAN} can be expressed through a two-player minimax game
\begin{center}
%\hspace{-0.1cm}
\includegraphics[width=\columnwidth]{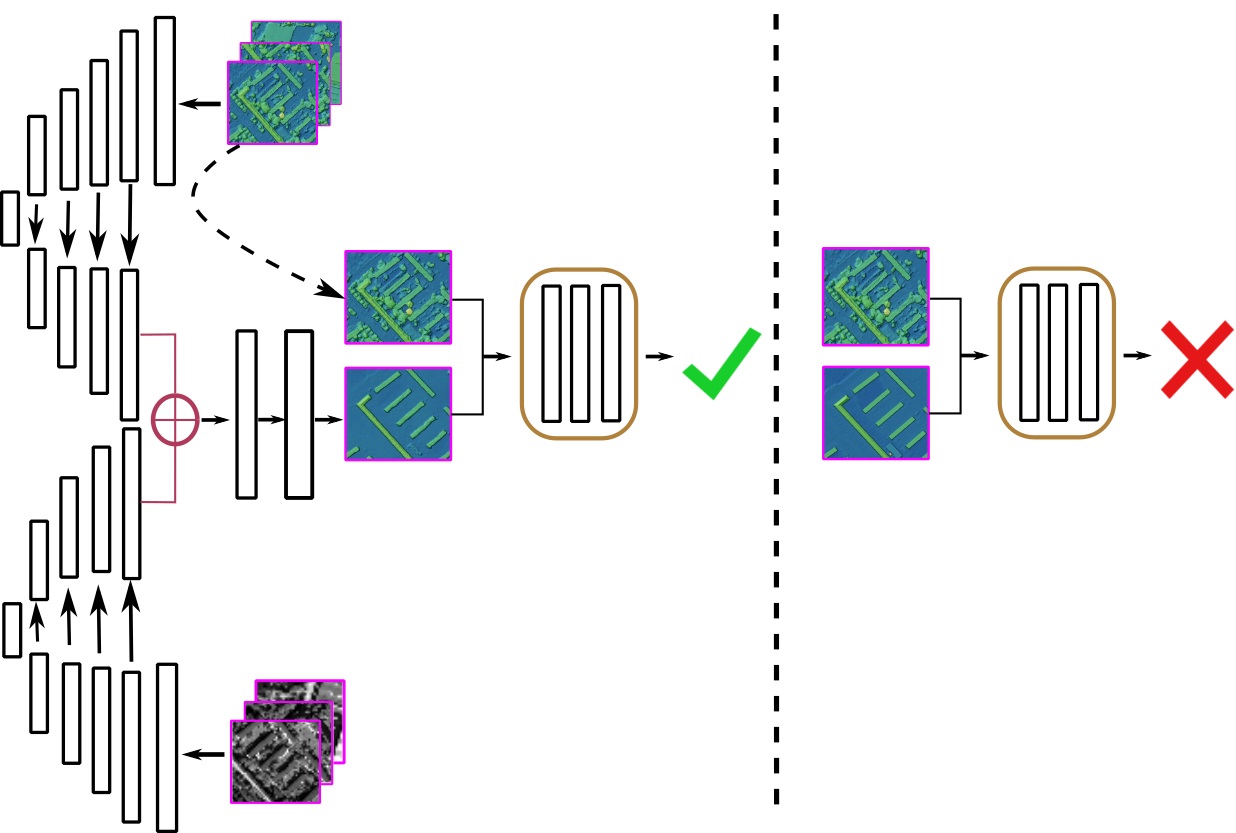}
\put (-205,166) {\scriptsize{stereo \gls{gl:DSM} ($I_1$)}}
\put (-200,-1) {\scriptsize{PAN} ($I_2$)}
\put (-133,116) {\scriptsize{\discriminator}}
\put (-243,81) {\scriptsize{\generator}}
\put (-180,67) {\scriptsize{\generator($I_1,I_2$)}}
\put (-168,118) {\scriptsize{$I_1$}}
\put (-73,118) {\scriptsize{$I_1$}}
\put (-77,67) {\scriptsize{GT}}
\put (-39,116) {\textbf{\scriptsize{\discriminator}}}
\put (-207,58) {\rotatebox{90}{\color{black}\textbf{\scriptsize{Fusion}}}}
\put (-187,70) {\rotatebox{90}{\color{black}\textbf{\scalebox{.5}{conv $1\times 1$}}}}
\captionof{figure}{\label{fig:architecture} Schematic overview of the proposed architecture for the building shape refinement in the
3D surface model by WNet-\gls{gl:cGAN} \space using depth and spectral information.}
\end{center}

\begin{equation}
\label{eq:objective_function}
G^\star = \arg \underset{G}{\min} \; \underset{D}{\max} \; \mathcal{L}_{\text{cGAN}}(G,D)+\lambda \mathcal{L}_{L_1}(G)
\end{equation}
between the generator and the discriminator, where \generator intents to minimize the objective function $\mathcal{L}_{\text{cGAN}}(G,D)$ against the \discriminator that aims to maximize it. 
Moreover, it should be mentioned that in the first term of \cref{eq:objective_function} we use an objective function with least squares
instead of the common negative log likelihood.
The second term in \cref{eq:objective_function} regularizes the generator and produces the output near the ground truth in a $L_1$ sense. 

In our previous work, we already adapted the architecture proposed by \citet{isola2016image} to obtain refined 3D surface models from the noisy and inaccurate stereo \glspl{gl:DSM}. 
Now, we propose a new \gls{gl:cGAN} architecture that integrates depth information from stereo \glspl{gl:DSM} together with spectral information from \gls{gl:PAN} images, as the latter provides a sharper information about building silhouettes, which allows not only a better reconstruction of building outlines but also their missing construction parts.
Since intensity and depth information have different physical meanings, we propose a hybrid \generator network where two separate \emph{UNet} \cite{ronneberger2015u} type of networks with the same architecture are used: we feed one part with the \gls{gl:PAN} image and the second part with the stereo \gls{gl:DSM} generating a so-called \emph{WNet} architecture. 
Before the last upsampling layer, which leads to the final output size, we concatenate the intermediate features from both streams.
Moreover, we increase the network with an additional convolutional layer of size $1 \times 1$, which plays the role of information fusion from different modalities.
As investigated earlier, this fusion can correct small failures in the predictions by automatically learning which stream of the network provides the best prediction result~\cite{bittner2018building}.
Finally, the $\operatorname{tanh}$ activation function $\sigma_\text{tanh}(z) = \operatorname{tanh}(z)$ is applied on the top layer of the \generator network.
\discriminator is represented by a binary classification network with a \emph{sigmoid} activation function $\sigma_\text{sigm}(z) = \frac{1}{1+\e^{-z}}$ to the top layer to output the probability that the input image belongs either to class 1 (\enquote{real}) or class 0 (\enquote{generated}).
It has five convolutional layers which are followed by a leaky \gls{gl:ReLU} activation function 
\begin{align*}
\sigma_\text{leaky ReLU}(z)=
	\begin{cases}
	  z, & \text{if $z>0$}\\
    az, & \text{otherwise}
   \end{cases}       
\end{align*}
%\begin{math}
%\sigma_\text{leaky ReLU}(z)=
 % \left\{
 %   \begin{array}{ll}
 %     z, & \text{if $z>0$}\\
 %     az, & \text{otherwise},
 %   \end{array}
 % \right.
%\end{math}
with a negative slope $a$ of 0.2.
The input to \discriminator is a concatenation of a stereo \gls{gl:DSM} with either a WNet-generated 3D surface model or a ground-truth \gls{gl:DSM}. 
A simplified representation of the proposed network architecture is demonstrated in \cref{fig:architecture}.

\section{Study Area and Experiments}
\label{sec:Experiments}

Experiments have been performed on WorldView-1 data showing the city of Berlin, Germany, within a total area of \SI{410}{\square\kilo\meter}. 
As input data, we used a stereo \gls{gl:DSM} and one of six very high-resolution \gls{gl:PAN} images, both with a resolution of \SI{0.5}{\meter}.
The \gls{gl:PAN} image is orthorectified.
As ground truth, the \gls{gl:LoD}2-\gls{gl:DSM}, generated with a resolution of \SI{0.5}{\meter} from a \gls{gl:CityGML} data model, was used for learning the mapping function between the noisy DSM and the \gls{gl:LoD}2-\gls{gl:DSM} with better building shape quality.
The detailed methodology on \gls{gl:LoD}2-\gls{gl:DSM} creation is given in our previous work.
A \gls{gl:CityGML} data model is freely available on the download portal Berlin 3D (\url{http://www.businesslocationcenter.de/downloadportal}).

\begin{figure*}[!t]
  \centering
	\subcaptionbox{\label{fig:OPatch_xs1910_ys23369}\scriptsize{\textbf{PAN}}}
                  {\hspace{0.27cm}\includegraphics[width=.17\linewidth]{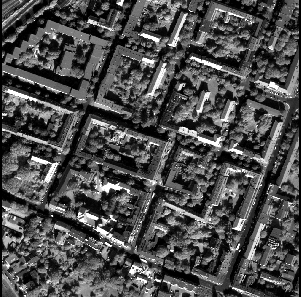}\hspace{0.2cm}}
  \subcaptionbox{\label{fig:DPatch_xs1910_ys23369_col}\scriptsize{\textbf{Stereo \gls{gl:DSM}}}}
                  {\hspace{0.27cm}\includegraphics[width=.17\linewidth]{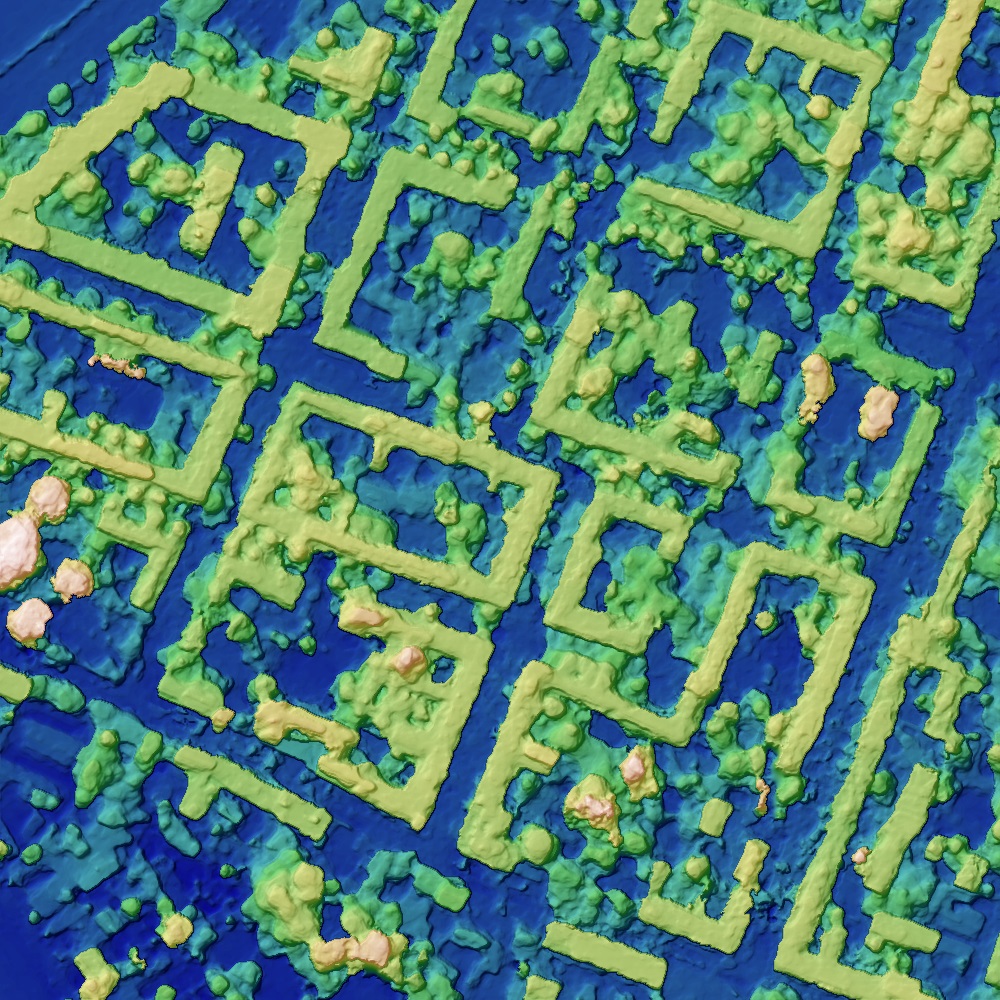}\hspace{0.2cm}}  
  \subcaptionbox{\label{fig:Patch_xs1910_ys23369_col}\scriptsize{\textbf{\gls{gl:cGAN} \gls{gl:LoD}2-\gls{gl:DSM}}}}
                  {\hspace{0.15cm}\includegraphics[width=.17\linewidth]{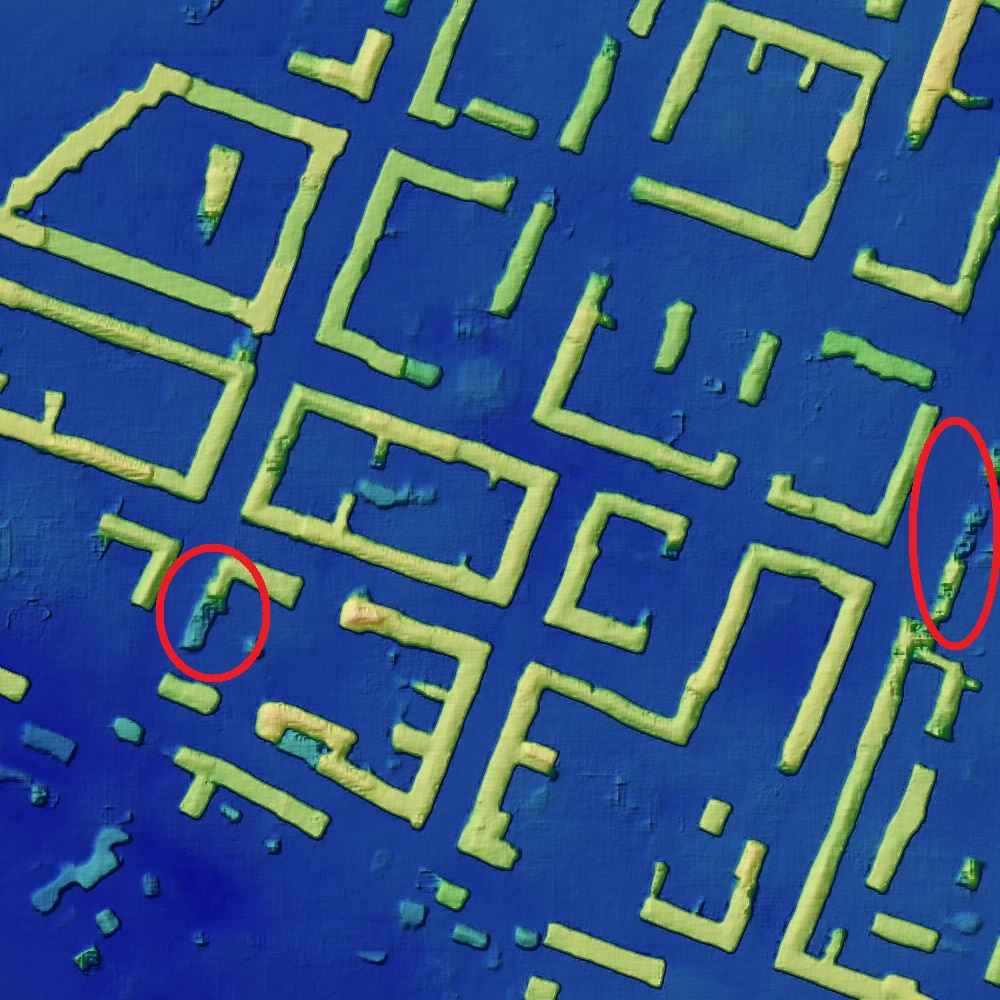}\hspace{0.2cm}}
  \subcaptionbox{\label{fig:FPatch_xs1910_ys23369_col}\scriptsize{\textbf{WNet-\gls{gl:cGAN} \gls{gl:LoD}2-\gls{gl:DSM}}}}
                  {\hspace{0.15cm}\includegraphics[width=.17\linewidth]{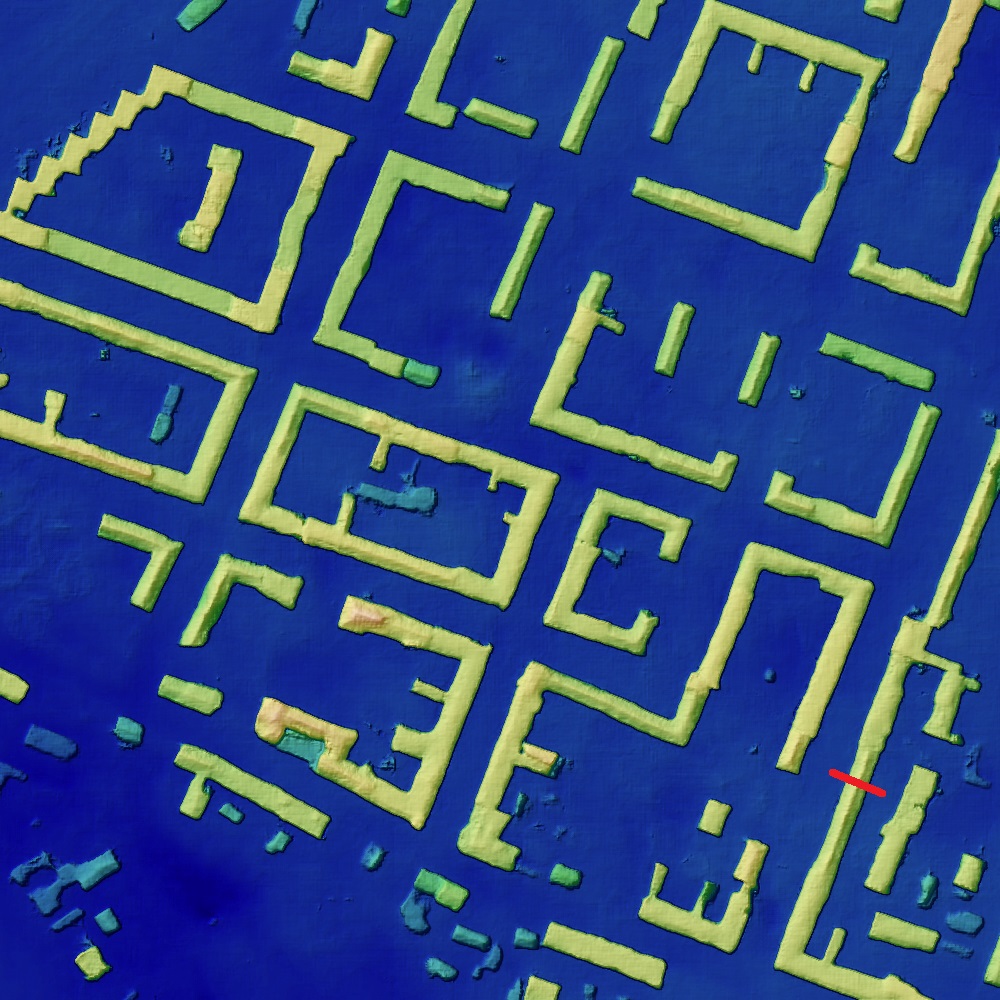}
									\put (1,14) {\color{red}\textbf{\scriptsize{1}}}
									%\put (1,57) {\color{red}\textbf{\scriptsize{2}}}
									%\put(-20,16){ \linethickness{2pt}\color{red}\line(3,-1){10}}
									%\tikz{\draw[red, thick, rotate=30] (3,1.5) -- ($(3,1.5)!1cm!(3,4.5)$);}
									}	
  \subcaptionbox{\label{fig:LODPatch_xs1910_ys23369_col}\scriptsize{\textbf{GT: \gls{gl:LoD}2-\gls{gl:DSM}}}}
                  {\hspace{0.15cm}\includegraphics[width=.17\linewidth]{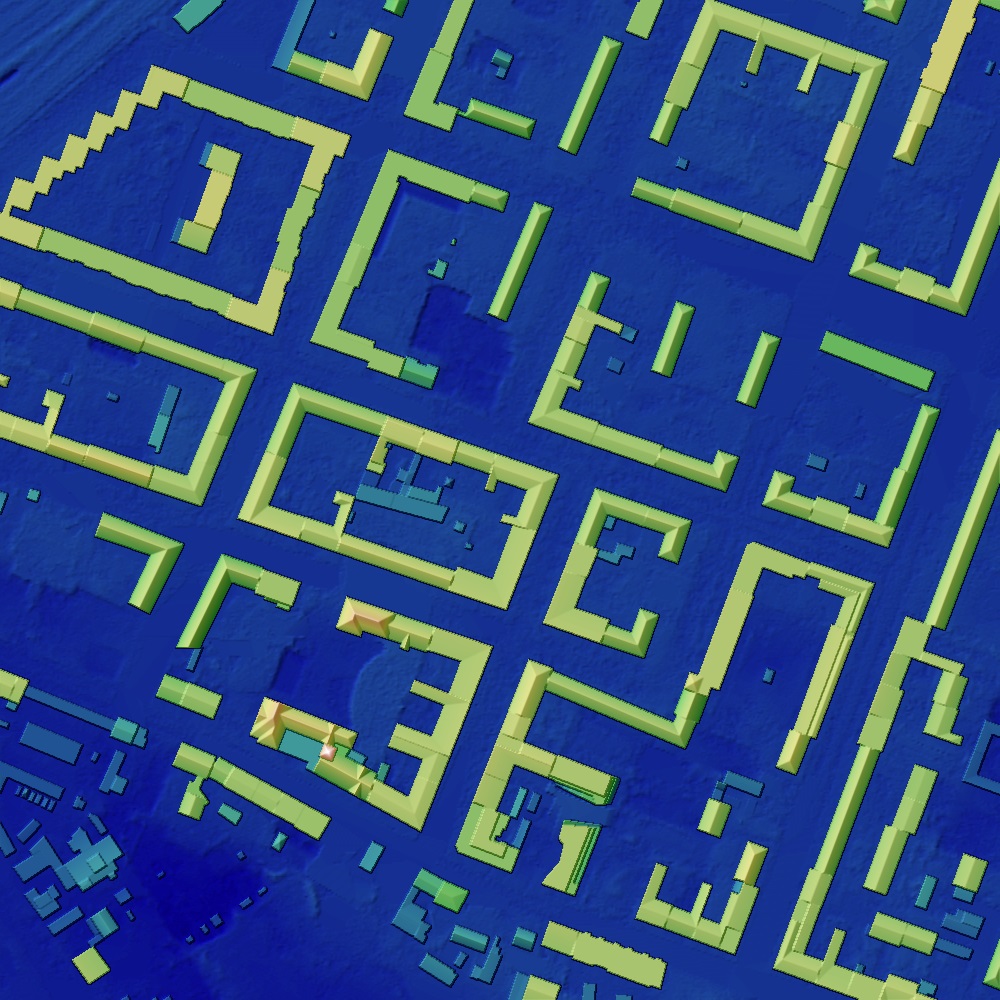}\hspace{0.2cm}} \\
	\vspace{0.1cm}
	% sample with central buildings 								
	\subcaptionbox{\label{fig:OPatch_xs4439_ys17996}\scriptsize{\textbf{PAN}}}
                  {\hspace{0.27cm}\includegraphics[width=.17\linewidth]{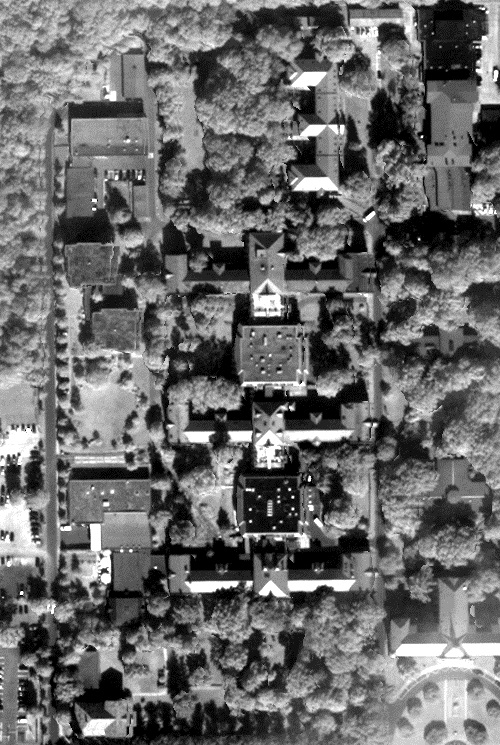}\hspace{0.2cm}}
  \subcaptionbox{\label{fig:DPatch_xs4439_ys17996_col}\scriptsize{\textbf{Stereo \gls{gl:DSM}}}}
                  {\hspace{0.27cm}\includegraphics[width=.17\linewidth]{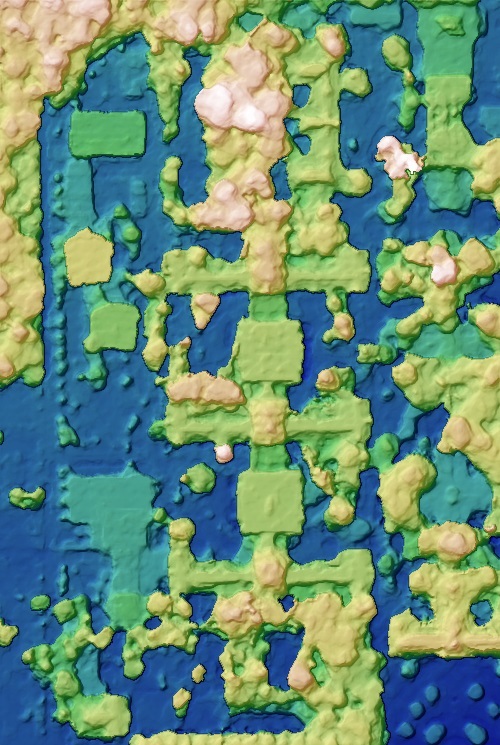}\hspace{0.2cm}}  
  \subcaptionbox{\label{fig:Patch_xs4439_ys17996_col}\scriptsize{\textbf{\gls{gl:cGAN} \gls{gl:LoD}2-\gls{gl:DSM}}}}
                  {\hspace{0.15cm}\includegraphics[width=.17\linewidth]{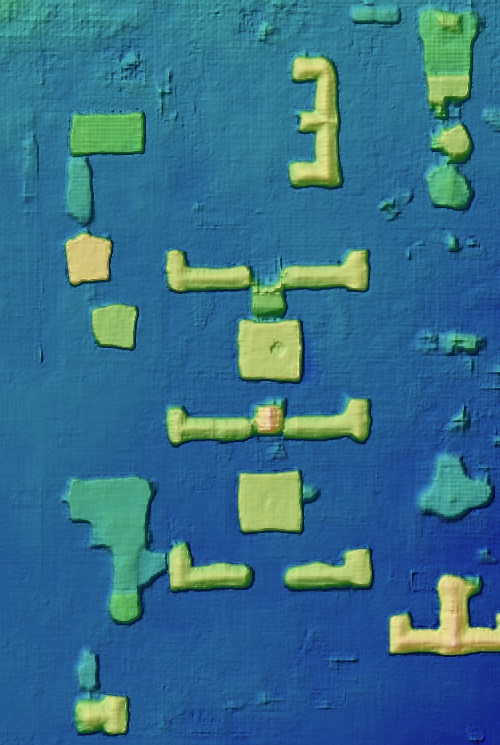}\hspace{0.2cm}}
  \subcaptionbox{\label{fig:FPatch_xs4439_ys17996_col}\scriptsize{\textbf{WNet-\gls{gl:cGAN} \gls{gl:LoD}2-\gls{gl:DSM}}}}
                  {\hspace{0.15cm}\includegraphics[width=.17\linewidth]{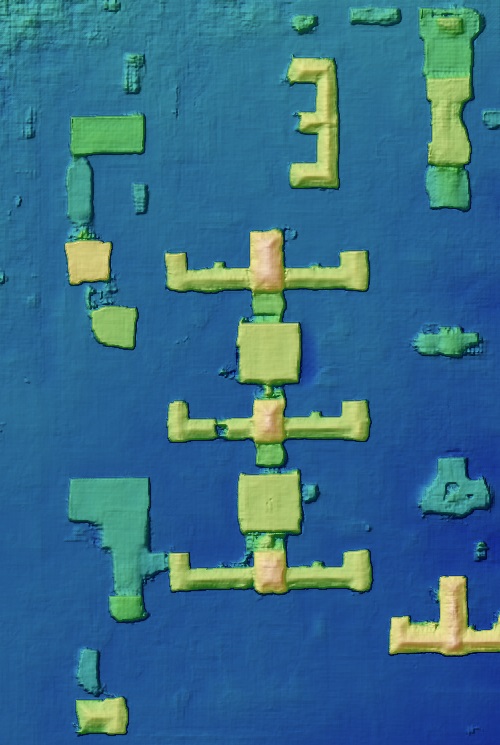}
									\put(-7,24){ \linethickness{0.7pt}\color{red}\line(-1,0){7}}
									\put (1,22) {\color{red}\textbf{\scriptsize{2}}}
									\put(-69,83){ \linethickness{0.7pt}\color{red}\line(-1,0){10}}
									\put (-90,81) {\color{red}\textbf{\scriptsize{3}}}
									\put(-38,84){ \linethickness{0.7pt}\color{red}\line(-1,0){10}}
									\put (1,82) {\color{red}\textbf{\scriptsize{4}}}
									}										
  \subcaptionbox{\label{fig:LODPatch_xs4439_ys17996_col}\scriptsize{\textbf{GT: \gls{gl:LoD}2-\gls{gl:DSM}}}}
                  {\hspace{0.15cm}\includegraphics[width=.17\linewidth]{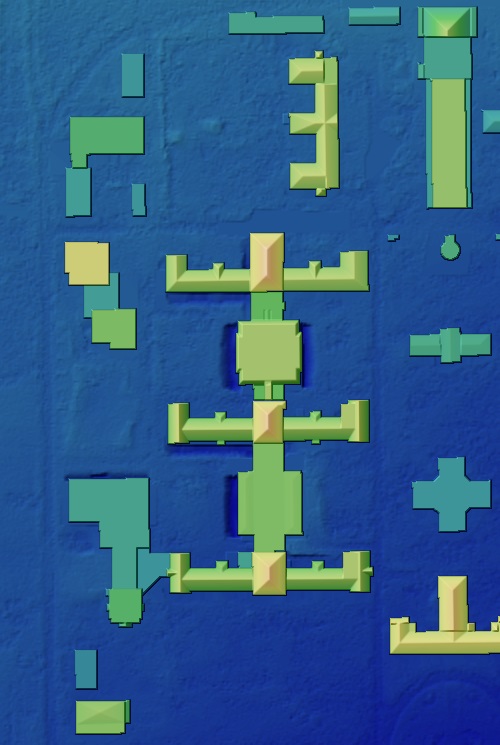}\hspace{0.2cm}} \\
	% profiles
	%\subcaptionbox{\label{fig:profile12}\scriptsize{\textbf{Profile 1}}}
                  %{\hspace{0.27cm}\includegraphics[width=.22\linewidth]{profile121}\hspace{0.2cm}}
  %\subcaptionbox{\label{fig:profile13}\scriptsize{\textbf{Profile 2}}}
                  %{\hspace{0.27cm}\includegraphics[width=.2\linewidth]{profile13}\hspace{0.2cm}}  
  %\subcaptionbox{\label{fig:profile11}\scriptsize{\textbf{Profile 3}}}
                  %{\hspace{0.15cm}\includegraphics[width=.22\linewidth]{profile111}\hspace{0.2cm}}
  %\subcaptionbox{\label{fig:profile14}\scriptsize{\textbf{Profile 4}}}
                  %{\hspace{0.15cm}\includegraphics[width=.22\linewidth]{profile14}}	
  %\subcaptionbox{\label{fig:profile15}\scriptsize{\textbf{Profile 5}}}
                  %{\hspace{0.15cm}\includegraphics[width=.22\linewidth]{profile15}\hspace{0.2cm}} \\
	%\vspace{-0.3cm}
	%\begin{center}
		%\tikz{\path[draw=black,fill=red] (0,0) rectangle (0.6cm,0.3cm);}\hspace{0.1cm}\text{\scriptsize{\textbf{Stereo \gls{gl:DSM}}}}\quad 
		%\tikz{\path[draw=black,fill=green] (0,0) rectangle (0.6cm,0.3cm);}\hspace{0.1cm}\text{\scriptsize{\textbf{GT: \gls{gl:LoD}2-\gls{gl:DSM}}}}\quad	
		%\tikz{\path[draw=black,fill=blue] (0,0) rectangle (0.6cm, 0.3cm);}\hspace{0.1cm}\text{\scriptsize{\textbf{\gls{gl:cGAN} \gls{gl:LoD}2-like \gls{gl:DSM}}}}\quad
		%\tikz{\path[draw=black,fill=magenta] (0,0) rectangle (0.6cm,	0.3cm);}\hspace{0.1cm}\text{\scriptsize{\textbf{WNet-\gls{gl:cGAN} \gls{gl:LoD}2-like \gls{gl:DSM}}}}
	%\end{center}
	\vspace{-0.3cm}
  \caption{Visual analysis of \glspl{gl:DSM}, generated by stereo \gls{gl:cGAN} and WNet-\gls{gl:cGAN} architectures, over selected urban areas. The \gls{gl:DSM} images are color-shaded for better visualization.\label{fig:Sample1}}
\end{figure*}
	
The implementation of the proposed WNet-\gls{gl:cGAN} is done with the \emph{PyTorch} python package.
For the training process, the satellite images were tiled into patches of size \SI{256x256}{\px} to fit into the single \textsc{NVIDIA TITAN X (Pascal) GPU} with \SI{12}{\giga\byte} of memory. 
The total number of epochs was set to 200 with a batch size of 5. 
We trained the \gls{gl:DSM}-to-\gls{gl:LoD}2 WNet-\gls{gl:cGAN} network with minibatch \gls{gl:SGD} using the ADAM optimizer.
An initial learning rate was set to $\alpha = 0.0002$ and the momentum parameters to $\beta_1 = 0.5$ and $\beta_2 = 0.999$. 

\section{Results and Discussion }
\label{sec:Results}
 
Two selected areas of the resulting \gls{gl:LoD}2-like \gls{gl:DSM} generated from combined spectral and depth information together with the \gls{gl:LoD}2-like \gls{gl:DSM} from a single image are illustrated in \cref{fig:Sample1}. 
From \cref{fig:DPatch_xs1910_ys23369_col} and \cref{fig:DPatch_xs4439_ys17996_col} we can see that the refinement of building
shapes only from stereo \glspl{gl:DSM} is a very challenging task, due to several reasons. 
First of all, the presence of vegetation can influence the reconstruction as some parts of buildings are covered by trees. 
Besides, the stereo \gls{gl:DSM} is very noisy itself, due to failures in the generation algorithms. 
It means that in most cases the types of roofs and, as a result, their shapes are indistinguishable. 
On the other hand, looking at \cref{fig:OPatch_xs1910_ys23369} and \cref{fig:OPatch_xs4439_ys17996} we can see that the edges and outlines can be seen very well in the \gls{gl:PAN} image.
Refinement of 3D buildings only from \gls{gl:PAN} image though would be very difficult as it does not contain 3D information,
which is very important. 
Therefore, the combination of these two types of information is a good compromise which leads to advantages.

It can be clearly seen that the hybrid WNet-\gls{gl:cGAN} architecture is able to reconstruct more complete building structures than the \gls{gl:cGAN} from a single data source (see the highlighted buildings in \cref{fig:FPatch_xs1910_ys23369_col}).
Even complicated forms of buildings are also preserved in the reconstructed 3D surface model.
The obvious example is a zigzag-shaped building at the upper-left part in \cref{fig:FPatch_xs1910_ys23369_col}.
This information could be only obtained from the \gls{gl:PAN} image (see \cref{fig:OPatch_xs1910_ys23369}).   

\begin{figure*}[!t]
  \centering
	% profiles
	\subcaptionbox{\label{fig:profile12}\scriptsize{\textbf{Profile 1}}}
                  {\hspace{0.27cm}\includegraphics[width=.22\linewidth]{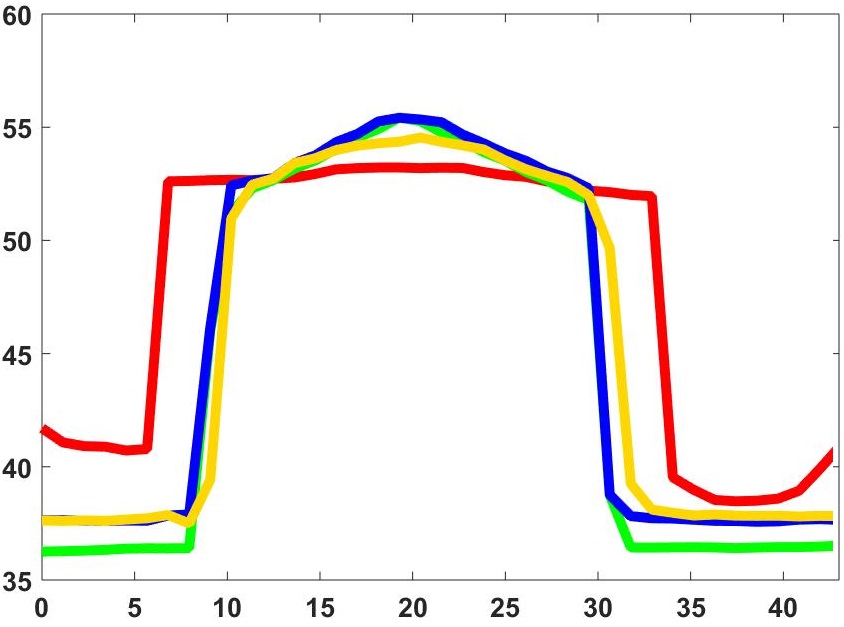}\hspace{0.2cm}}
  %\subcaptionbox{\label{fig:profile13}\scriptsize{\textbf{Profile 2}}}
                  %{\hspace{0.27cm}\includegraphics[width=.2\linewidth]{profile13}\hspace{0.2cm}}  
  \subcaptionbox{\label{fig:profile11}\scriptsize{\textbf{Profile 3}}}
                  {\hspace{0.15cm}\includegraphics[width=.22\linewidth]{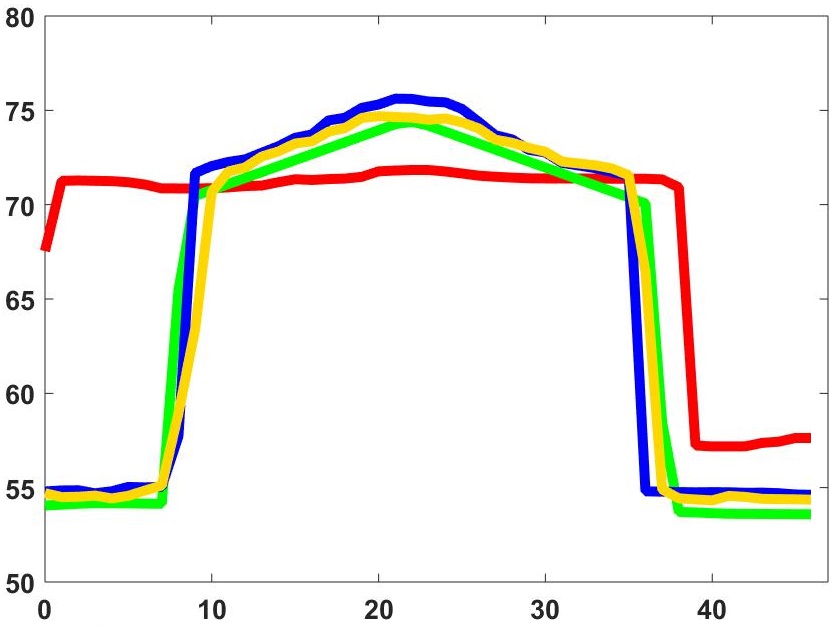}\hspace{0.2cm}}
  \subcaptionbox{\label{fig:profile14}\scriptsize{\textbf{Profile 4}}}
                  {\hspace{0.15cm}\includegraphics[width=.22\linewidth]{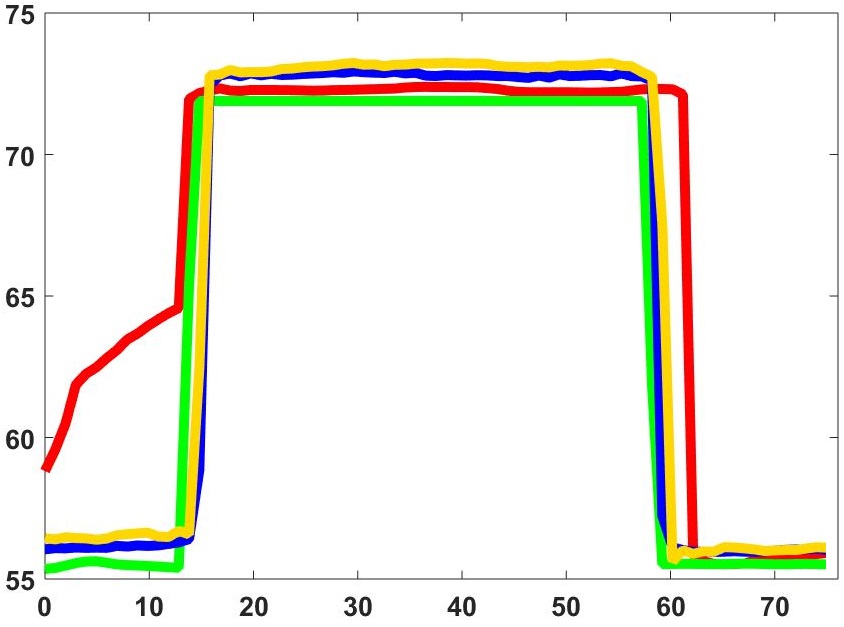}}	
  \subcaptionbox{\label{fig:profile15}\scriptsize{\textbf{Profile 5}}}
                  {\hspace{0.15cm}\includegraphics[width=.22\linewidth]{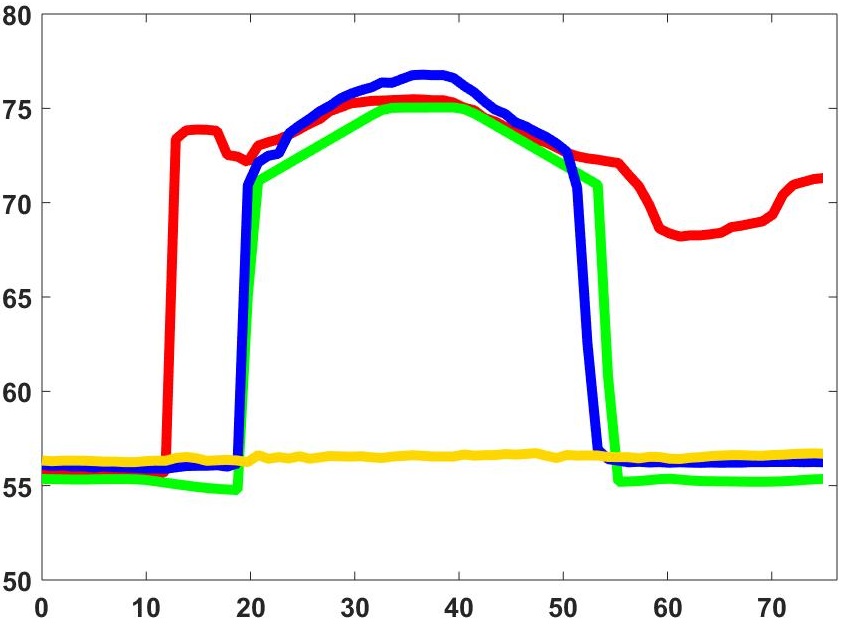}\hspace{0.2cm}} \\
	\vspace{-0.3cm}
	\begin{center}
		\tikz{\path[draw=black,fill=red] (0,0) rectangle (0.6cm,0.3cm);}\hspace{0.1cm}\text{\scriptsize{\textbf{Stereo \gls{gl:DSM}}}}\quad 
		\tikz{\path[draw=black,fill=green] (0,0) rectangle (0.6cm,0.3cm);}\hspace{0.1cm}\text{\scriptsize{\textbf{GT: \gls{gl:LoD}2-\gls{gl:DSM}}}}\quad	
		\tikz{\path[draw=black,fill=yellow] (0,0) rectangle (0.6cm, 0.3cm);}\hspace{0.1cm}\text{\scriptsize{\textbf{\gls{gl:cGAN} \gls{gl:LoD}2-like \gls{gl:DSM}}}}\quad
		\tikz{\path[draw=black,fill=blue] (0,0) rectangle (0.6cm,	0.3cm);}\hspace{0.1cm}\text{\scriptsize{\textbf{WNet-\gls{gl:cGAN} \gls{gl:LoD}2-like \gls{gl:DSM}}}}
	\end{center}
	\vspace{-0.3cm}
  \caption{Visual analysis of selected building profiles in generated \glspl{gl:DSM}.}
\end{figure*}

The second example depicts a smaller but scaled area for better visual investigation. 
Here, a central building is complete and more details are distinguishable. 
Besides, the ridge lines of the roofs are also much better visible. 
One can even guess to which type of roof parts of building belong to: gable or hip roofs. 
A clear contribution of the spectral information to the building shape refinement task can be seen at the upper-right building in \cref{fig:FPatch_xs4439_ys17996_col}. 
We can notice that this building structure is more complete. 
The outlines of all buildings are clearer rectilinear and the building shapes become more symmetrical. 
To look more detailed into the 3D information, we illustrate some building profiles. 
We can see that the roof forms like gable and hip are clearly improved. 
The ridge lines tend to be sharp peaks. 
With the profile in \cref{fig:profile15} we again highlight the ability of the proposed architecture to reconstruct even complicated buildings, which is difficult to reconstruct using a single stereo \gls{gl:DSM} information.  

\begin{table}[!b]
%\scriptsize 
\footnotesize
\caption{Prediction accuracies of \gls{gl:cGAN} and Fused-\gls{gl:cGAN} models on investigated metrics over the Berlin area.}\label{tab:eval_results} %Binary classification results on the test dataset in comparison to the four-layer network.
\centering
\begin{tabular}{|c|c|c|c|c|c|} %{p{0.25\linewidth}p{0.25\linewidth}p{0.25\linewidth}}
\hline
& \textbf{MAE, \si{\metre}} & \textbf{RMSE, \si{\metre}} & \textbf{NMAD, \si{\metre}} & \textbf{NCC, \si{\metre}}\\ 
\hline
\textbf{Stereo \gls{gl:DSM}}& 3.00 & 5.97 & 1.48 & 0.90 \\ 
\textbf{\gls{gl:cGAN}} & 2.01 & 4.78 & 0.86 & 0.92 \\ 
\textbf{Fused-\gls{gl:cGAN}} & \textbf{1.79} & \textbf{4.36} & \textbf{0.67} & \textbf{0.94} \\
%\hline
\hline
\end{tabular}
\end{table}

To quantify the quality of the generated \glspl{gl:DSM}, we evaluated the metrics \gls{gl:MAE}, \gls{gl:RMSE}, \gls{gl:NMAD} and \gls{gl:NCC}, commonly used for 3D surface model accuracy investigation, on the \gls{gl:cGAN} and WNet-\gls{gl:cGAN} setups and report their performance in \cref{tab:eval_results}. 
As we are interested in quantifying the improvements only of the building shapes on \glspl{gl:DSM} the above mentioned metrics were measured only within the area where buildings are present plus a three-pixel buffer around each of them. 
This was achieved by employing the binary building mask and dilation procedure on the footprint boundaries. 
From the obtained results we can see that \gls{gl:DSM} from WNet-\gls{gl:cGAN} is better than the original stereo \gls{gl:DSM} and the \gls{gl:DSM} generated by the \gls{gl:cGAN} model on all proposed metrics.
This is reasonable, as the spectral information provides additional information, helpful to reconstruct building structure more accurately and detailed, which is not possible using only stereo DSM.
This feature especially influences the corners, outlines, and ridge lines. 
As the \gls{gl:NCC} metric indicates how the form of the object resembles the ground truth object, the gaining of 4 \% in comparison to the stereo \gls{gl:DSM} and 2 \% improvement on the \gls{gl:DSM} generated by \gls{gl:cGAN} model over the whole test area, which includes thousands of buildings, demonstrate the advantage of using complementary information for such complicated tasks.
The high values of \gls{gl:RMSE} (order of \SI{5}{\meter}) is due to data acquisition time difference between the available DSM generated from stereo satellite images and the given ground truth data.
As a result, several buildings are not presented or newly constructed in the more recent data set. 

%as the metric evaluation between all buildings which are present on generated \gls{gl:DSM} but not presented on reference \gls{gl:LoD}2-\gls{gl:DSM} or vice versa. 
%This is due to data acquisition time difference between the available DSM generated from stereo satellite images and the given ground truth data.

%%%%%%%%%%%%%%%%%%%%%%%%%%%%%%%%%%%%%%%%%%
\glsresetall
\section{Conclusion}\label{sec:conclusion}

Refinement and filtering techniques from the literature for \glspl{gl:DSM} quality improvement are adequate for either 
small-scale \glspl{gl:DSM} or \glspl{gl:DSM} with no discontinuities. 
As a result, there is a need to develop a refinement procedure that can handle discontinuities, mainly building forms in urban regions, in high-resolution large-scale \glspl{gl:DSM}.
A common strategy in remote sensing for refinement procedures is the use of all available information from different data sources.
Their combination helps to compensate the mistakes and gaps in each independent data source.

We present a method for automatic large-scale \gls{gl:DSM} generation with refined building shapes to the \gls{gl:LoD} 2 from multiple spaceborne remote sensing data on the basis of \glspl{gl:cGAN}.
The designed end-to-end WNet-\glspl{gl:cGAN} integrates the contextual information from height and spectral images to produce good-quality 3D surface models.
The obtained results show the potential of the proposed methodology to generate more completed building structures in \glspl{gl:DSM}.
The network is able to learn how to complement the strong and weak sides of \gls{gl:PAN} image and stereo \gls{gl:DSM}, as, for
instance, the stereo \glspl{gl:DSM} provide elevation information of the objects, but \gls{gl:PAN} images provide texture information and, as a result, more accurate building boundaries and silhouettes.

% To start a new column (but not a new page) and help balance the last-page
% column length use \vfill\pagebreak.
% -------------------------------------------------------------------------
\vfill
\pagebreak

% References should be produced using the bibtex program from suitable
% BiBTeX files (here: strings, refs, manuals). The IEEEbib.bst bibliography
% style file from IEEE produces unsorted bibliography list.
% -------------------------------------------------------------------------
%\bibliographystyle{IEEEbib}
%\bibliography{refs}

\printbibliography

\end{document}